%% file: main.tex
\documentclass[letterpaper,10pt,conference]{ieeeconf}
\IEEEoverridecommandlockouts
\usepackage{amsmath,amssymb}
\usepackage{mathptmx}
\usepackage[protrusion=false]{microtype}
\usepackage{graphicx}
\usepackage{capt-of}
\usepackage{cite}
\usepackage{url}
\usepackage{booktabs,tabularx,array}
\usepackage{tikz}
\usetikzlibrary{arrows.meta,positioning,fit,backgrounds,calc}
\usepackage{balance}
\newcommand{\clip}{\operatorname{clip}}
\newcommand{\E}{\mathbb{E}}
\newcolumntype{Y}{>{\raggedright\arraybackslash}X}
\title{\LARGE\bf Terrain-Indexed Motion Priors and Geometric Support\\
for Extreme Humanoid Parkour}
\author{
  Chenghao Sun$^{1*}$, Youhao Lyu$^{2*}$, Zizheng Li$^{3}$, Shuting Zhang$^{4}$, Ziyang Dong$^{1}$, and Wenhao Cui$^{1\dagger}$%
  \thanks{*Equal contribution.}%
  \thanks{$^{\dagger}$Corresponding author: Wenhao Cui.}%
  \thanks{$^{1}$Chenghao Sun, Ziyang Dong, and Wenhao Cui are with Noetix Robotics, Beijing, China.}%
  \thanks{$^{2}$Youhao Lyu is with the University of Science and Technology of China (USTC), Hefei, China.}%
  \thanks{$^{3}$Zizheng Li is with [Institution Name], [City], [Country].}%
  \thanks{$^{4}$Shuting Zhang is with the National University of Defense Technology (NUDT), Changsha, China.}%
}
\IEEEaftertitletext{%
  \begin{minipage}{\textwidth}
  \centering
  \includegraphics[width=\textwidth,trim=118bp 227bp 180bp 171bp,clip]{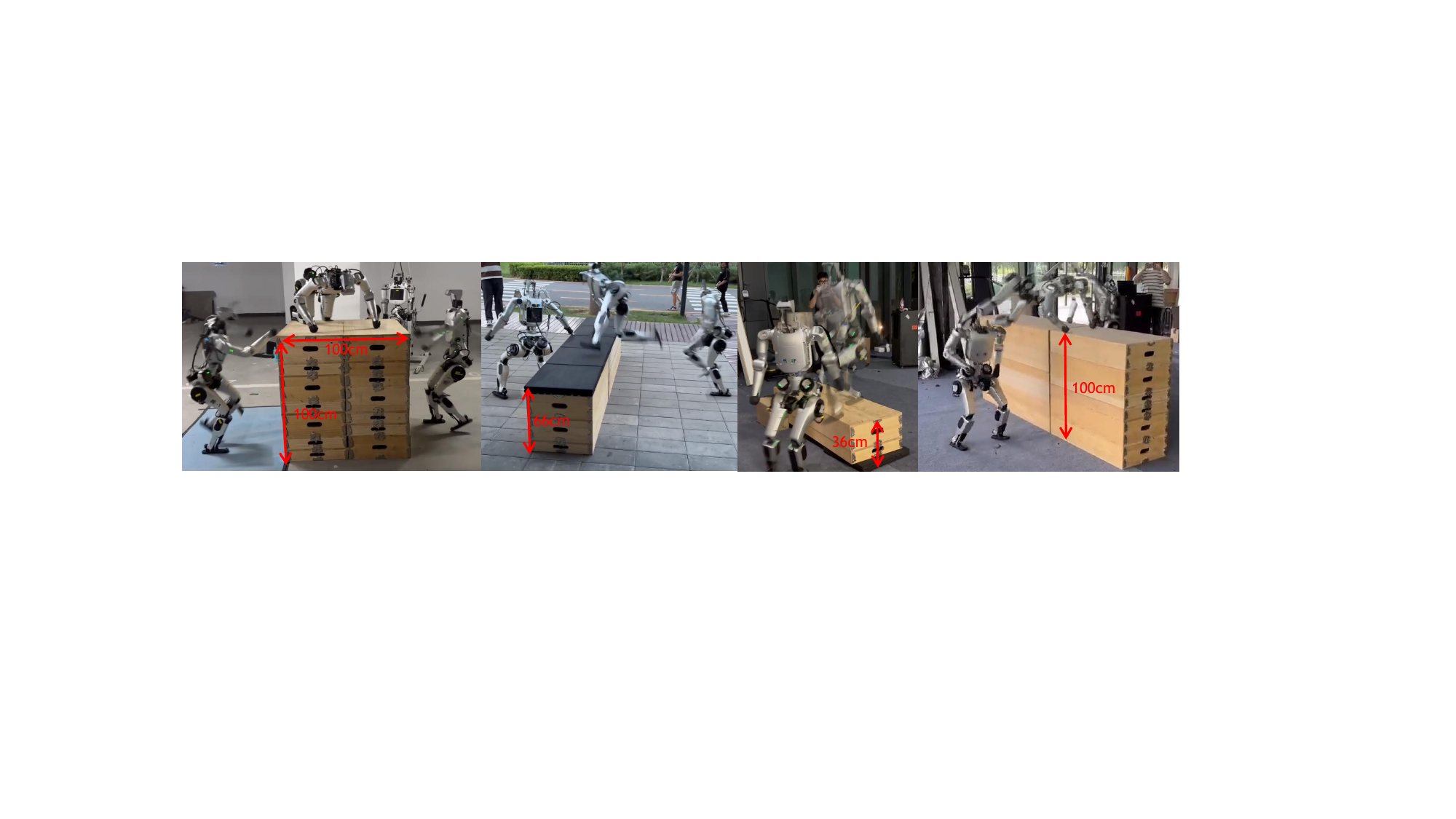}
  \captionof{figure}{Our framework enables the Noetix E1 humanoid to
  autonomously select terrain-appropriate skills using onboard perception
  and traverse diverse obstacle types, with generalization to the tested
  terrain heights outside the training ranges.}
  \label{fig:cover}
  \end{minipage}\vspace{0.6em}%
}
\begin{document}
\bstctlcite{IEEE:BSTcontrol}
\maketitle
\thispagestyle{empty}
\pagestyle{empty}

\begin{abstract}
We present a progressive imitation framework that learns terrain-appropriate
humanoid parkour skills and subsequently adapts their hand--knee support.
Human demonstrations provide traversal behaviors, but their contact patterns
need not suit a robot's proportions or the available support surfaces.
Our framework changes the role of demonstrations across three stages:
privileged imitation produces physically executable robot references;
terrain-indexed adversarial motion priors guide skill acquisition; and
geometric support guidance with a reduced motion-prior weight enables
contact adaptation. Candidate-based precontact attraction and actual-contact
evaluation guide support without prescribing a contact sequence.
The deployed policy selects behaviors from onboard depth, proprioception,
and velocity commands without an external skill label.
On the Noetix E1, the policy achieves 96.34\% hybrid success in simulation
and transfers without real-world policy fine-tuning, completing 29 of 30
individual obstacle trials and four of five hybrid courses.
Support-adaptation ablations show higher traversal success for the complete
adaptation stage on the tested terrains, while qualitative observations
show changes in contact placement. These results support acquiring skills
from demonstrations before relaxing their influence to accommodate
robot-specific support configurations.
\end{abstract}

\section{Introduction}
Humanoid parkour extends mobility beyond conventional walking through
whole-body obstacle traversal. These maneuvers require
coordinating whole-body contacts as the robot mounts, crosses, and leaves
an obstacle. Terrain geometry determines both the appropriate motion and
available support. Human demonstrations provide examples of
such coordination, but differences in body proportions, actuation, and
available support surfaces make direct reproduction an incomplete solution.

Learning-based control has enabled robust perceptive quadrupedal locomotion
in natural environments~\cite{miki_robustperceptive_2022} and diverse dynamic
bipedal skills~\cite{li_versatilebipedal_2025}.
Perceptive control has advanced humanoid obstacle traversal through visual
reinforcement learning and the use of human motion
information~\cite{zhuang_humanoidparkour_2024,wang_apexlearning_2026,wu_perceptivehumanoid_2026}.
Adversarial motion priors (AMP) provide behavioral guidance alongside task
rewards~\cite{peng_amp}, while multiple and terrain-conditioned priors organize
this guidance across skills and
environments~\cite{vollenweider_advancedskills_2022,huang_learningmultiskill_2025,sun_rawpixels_2026,guo_tgmp_2026}.
However, acquiring an executable skill does not alone determine
where it is appropriate or how its contacts should adapt. A demonstrated
contact pattern may be unsuitable for the robot or a different support layout.
We therefore ask: \emph{How can demonstrations guide the acquisition
of terrain-appropriate traversal skills while leaving the robot freedom to
adapt its support configuration?}

Three connected difficulties arise: mapped motions may violate robot
physics; executable references may be inappropriate for the encountered
terrain; and demonstrated contacts may restrict alternative support
configurations. Addressing physical executability therefore does not eliminate
the need for terrain-dependent guidance or subsequent contact adaptation. Our design
principle is to use imitation as a scaffold for skill acquisition and
subsequently relax the motion prior while increasing geometric support guidance.
The aim is
to preserve what the skill accomplishes while allowing its physical
realization to change.

We propose a \emph{progressive physics-grounded imitation framework} connecting
reference preparation, terrain-dependent skill learning, and support adaptation.
Privileged mimic training first prepares robot reference rollouts from human
motions. Terrain-organized reference pools then supervise adversarial motion
priors with a shared representation and separate terrain-specific heads,
guiding a single traversal policy. Support adaptation adds geometric hand--knee
candidates, precontact guidance, and actual-contact rewards, increasing support
weight while gradually reducing AMP weight to half its initial value to allow
robot-specific contact configurations rather than a prescribed sequence.
Terrain labels, reference discrimination, and support geometry are
confined to training. At deployment, the policy uses onboard depth, proprioceptive history,
and velocity commands without a separate skill command. The policy achieves 96.34\% hybrid traversal success in simulation. Our evaluation examines terrain-organized supervision and support adaptation.

The main contributions are:
\begin{itemize}
\item A progressive imitation framework that uses terrain-indexed motion
priors for skill acquisition and then relaxes them for support adaptation.
\item Geometric hand--knee support guidance combining precontact attraction
and actual-contact evaluation without prescribing a contact sequence.
\item Simulation comparisons and onboard hardware trials evaluating
single-policy traversal, support adaptation, and transfer to the tested
terrain-height shifts.
\end{itemize}

\input{figures/framework}

\section{Related Work}
\subsection{Perceptive Whole-Body Traversal}
Perceptive traversal requires both skill acquisition and integration.
Robot Parkour Learning distills quadrupedal obstacle skills into a single
vision-based policy without reference motions~\cite{zhuang_robotparkour_2023}.
Humanoid Parkour Learning learns visual whole-body obstacle skills without
motion priors~\cite{zhuang_humanoidparkour_2024}, while Deep Whole-body Parkour
integrates exteroception into multi-motion whole-body tracking on varied
terrain~\cite{zhuang_deepwholebody_2026}.
BeyondMimic combines humanoid motion tracking with guided diffusion for
flexible skill composition~\cite{liao_beyondmimic_2025}. ASAP uses real-world
data to learn a residual action model for sim-to-real dynamics
alignment~\cite{he_asap_2025}.
With motion references, Zhang et al. combine terrain-aware diffusion motion generation with
reinforcement-learning-based tracking~\cite{zhang_wholebodygeneration_2026},
while Perceptive Humanoid Parkour composes references through motion matching
and distills privileged tracking experts into a depth-conditioned
student~\cite{wu_perceptivehumanoid_2026}.
For simulated characters, HIL jointly uses motion tracking and adversarial
imitation to compose parkour skills learned from videos~\cite{wang_hil_2025}.
Multi-skill integration is also central to APEX, which learns contact-rich
behaviors with a reference-free progress reward and combines them by policy distillation~\cite{wang_apexlearning_2026},
and Light-Loco-Parkour, which combines expert distillation with reward-driven
skill transitions~\cite{chen_lightlocoparkour_2026}.
We combine motion priors with geometric support guidance to retain useful
reference behaviors while accounting for support-surface constraints,
to enable support adaptation after skill acquisition.

\subsection{Multiple and Conditional Motion Priors}
Motion priors differ in reference organization, conditioning, and parameter
sharing. AMP learns an adversarial motion reward from unstructured
examples~\cite{peng_amp}. Multiple AMP separates style objectives through
distinct discriminators, reference data, and replay
buffers~\cite{vollenweider_advancedskills_2022}.
Conditional AMP instead uses skill-conditioned discrimination and a skill
recognition objective~\cite{huang_learningmultiskill_2025}.
Beyond adversarial formulations, SMP derives reusable, task-agnostic motion
rewards from pre-trained diffusion models using score distillation sampling
for physics-based character control~\cite{mu_smp_2026}.
For terrain-dependent learning, Now You See That uses terrain-specific
reference datasets and discriminators, alongside critics with a shared
backbone and terrain-specific output heads~\cite{sun_rawpixels_2026}.
T-GMP learns a terrain-conditioned generative motion prior with adversarial
and foothold objectives~\cite{guo_tgmp_2026}.
We extend AMP's kinematic representation with torque, contact, and height-scan
channels. These inputs provide dynamics-related cues and local terrain
context to support learning motion--terrain compatibility beyond
kinematic similarity alone.

\subsection{Contact Priors for Robot Locomotion}
Geometry can guide support through policy inputs or rewards, before or
at contact.
PUMA provides the actor with predicted egocentric foothold distances and
headings alongside foothold-related rewards~\cite{wang_pumaperceptiondriven_2026}.
BeamDojo penalizes unsupported foot-sole regions at
contact~\cite{wang_beamdojolearning_2025}, whereas CReF extracts planar
foothold candidates and rewards touchdown proximity to the nearest
candidate~\cite{hao_crefcrossmodal_2026a}.
SSR provides dense swing-phase guidance by assessing support at predicted
landing positions and combines it with terrain-specific motion
priors~\cite{yu_ssr_2026}.
Beyond foot placement, whole-body maneuvers require contact locations tailored
to the distinct support requirements of hands and knees. Our geometric
support candidates guide these limbs toward suitable surface regions before
contact, while actual-contact evaluation assesses established support.

\section{Method}
\subsection{Framework Overview}
\label{sec:framework_overview}
Fig.~\ref{fig:pipeline} presents an overview of our asymmetric actor--critic
framework for perceptive humanoid parkour.
A separately optimized history estimator encodes proprioceptive observations
into a velocity estimate and a latent representation, while a visual encoder
extracts terrain features from onboard depth images.
The actor combines these features with current proprioception and velocity
commands through four gated experts to produce joint position targets.
During training, terrain-indexed AMP supplies terrain-matched style rewards
for skill acquisition, and geometric hand--knee support guidance is introduced
alongside prior relaxation for contact adaptation.
The critic uses privileged information during training, while the deployed
policy selects traversal behaviors from onboard observations without
terrain or skill-label inputs.

\textbf{Progressive reference use.}
Stage~1 (reference preparation) uses privileged mimic training to produce physically executable
robot rollouts with torque, contact, and height scans. Stage~2 (skill
acquisition) learns through terrain-indexed AMP (Sec.~\ref{sec:terrain_amp}).
Stage~3 (support adaptation) increases geometric support guidance and reduces
AMP weight to half its initial value (Sec.~\ref{sec:support_candidate}),
allowing alternative contacts.

\textbf{Learning objective.}
The policy reward combines task terms, geometric support, and a routed
motion prior:
\begin{equation}
 r_t=0.5\Delta t\,[R_{\rm other}(s_t,a_t)+\alpha(n)S_t]
       +\beta(n)\phi_k(U_t).
 \label{eq:reward}
\end{equation}
Here $s_t,a_t$ denote state/action, $k$ the terrain class, $U_t$ a motion
window, and $n$ the control-step index. The base task reward follows the
reward design of \texttt{legged\_gym}~\cite{rudin_learningtowalk_2022},
combining velocity tracking, action/energy regularization, and contact
penalties; $R_{\rm other}$ also includes inherited geometric terms.
$S_t$ combines support placement, retention, and penalties and is enabled
only in Stage~3; support-candidate rewards are inactive in Stages~1 and~2.
With the existing $0.5$ blend, reducing nominal AMP weight from two to one
corresponds to $\beta:1\rightarrow0.5$. The support weight $\alpha$ increases;
schedule details appear in Sec.~\ref{sec:support_candidate}.

\textbf{Policy and information boundary.}
The actor observes a 78-dimensional proprioception/command vector and a
$24\!\times\!24$ depth image clipped to $0.3$--$4$ m. A history estimator
(HIM) encodes five proprioceptive frames into a three-dimensional velocity
estimate and a normalized 16-dimensional latent. Together with 32 depth
features, these give a 129-dimensional actor input. Four gated actor experts
are distinct from the four discriminator heads. The critic has four experts
and receives privileged four-dimensional terrain labels, without a fixed
terrain-to-expert assignment or auxiliary label-classification loss.
PPO~\cite{schulman_proximalpolicy_2017} updates the actor, critic,
and depth encoder, while HIM uses a separate optimizer with privileged
velocity and next-observation representation targets.
True terrain labels, discriminator height scans, and support geometry remain in
training.

\subsection{Terrain-Indexed AMP}
\label{sec:terrain_amp}
\textbf{Design rationale.}
A shared motion-feature trunk and four terrain-specific output heads
represent terrain-indexed motion priors. Each head distinguishes policy
windows from a matching reference pool, and the current terrain selects
the training reward. Standing and walking references occur in multiple
pools, allowing shared behavior to coexist with terrain-specific traversal.
This organization is intended to reduce competition between common
locomotion and less frequent traversal skills. The deployed actor receives
no terrain label. Comparisons with Single AMP evaluate the complete
terrain-indexed formulation; they do not isolate trunk sharing or establish
that mode collapse is prevented.

\textbf{Enhanced motion representation.}
We augment kinematics with contact patterns, torque-based actuation cues,
and height scans describing local terrain. Together these channels provide
dynamics-related and motion--terrain information beyond movement similarity:
\begin{equation}
 x_t=[q_{23},p_{72},v_3,\omega_3,\dot q_{23},c_{24},\tau_{23},h_{231}]
 \in\mathbb{R}^{402}.
 \label{eq:amp_input}
\end{equation}
Here $q,\dot q$ are joint positions and velocities; $p$ contains 24 local body
positions; $v,\omega$ are root velocities; and $c,\tau,h$ denote contact,
torque, and yaw-aligned height-scan channels. Removing global root position
and quaternion from each 409-entry export yields this input. We validate the exported data for dimensional consistency, finite values, temporal consistency, and joint/body ordering.

\textbf{Terrain-indexed sampling.}
Table~\ref{tab:terrains} groups eight clips into four pools sharing standing
and forward motions across heights. Sampling weights are $0.5$ for standing,
turning, lateral, and backward clips and $3$ for forward/traversal clips,
normalized within each pool. Training uses terrain-eligible reference-state initialization with
probability $0.8$, whereas evaluation disables it. Expert windows
come from reference pools and policy windows from terrain replay slots.
Terrain labels select training priors, not deployed skill commands.

\begin{table}[t]
\caption{Reference pools. S/W: standing/forward walking; T/L/B: turning/lateral/
backward walking. Low/Mid/High label terrain-associated traversal references.}
\label{tab:terrains}\centering\small
\begin{tabular}{clcc}
\toprule
Head & Class & Training height (m) & Reference pool\\
\midrule
0 & High & $0.95$--$1.05$ & S, W, High\\
1 & Mid & $0.55$--$0.65$ & S, W, Mid\\
2 & Low & $0.35$--$0.45$ & S, W, Low\\
3 & Flat & 0 & S, T, L, B, W\\
\bottomrule
\end{tabular}
\end{table}

\textbf{Shared encoding and terrain heads.}
A shared empirical normalizer processes 402-dimensional frames. A
$231\!\rightarrow\!64\!\rightarrow\!32\!\rightarrow\!16$ scan encoder with LeakyReLU
hidden activations joins 171 other channels per frame. Four frames yield
$E_\theta(U_t)\in\mathbb R^{748}$ from normalized raw windows
$U_t\in\mathbb R^{1608}$. The discriminators are
\begin{equation}
 z_t=f_\theta(E_\theta(U_t)),\qquad D_k(U_t)=w_k^\top z_t+b_k,
 \label{eq:shared}
\end{equation}
where $f_\theta$ has widths 512 and 256 and $k\in\{0,1,2,3\}$ is the terrain
class. The scan encoder and trunk are shared; the four $256\!\rightarrow\!1$
heads are separate. The terrain label routes each sample to its corresponding discriminator head.

\textbf{Adversarial learning and reward.}
For every active class, expert windows $U^E$ come from its reference pool and
policy windows $U^\pi$ come from its replay slot. The implemented loss is
\begin{align}
 \mathcal L_k={}&\tfrac12\big[-\E\tanh(0.3D_k(U^E))
                  +\E\tanh(0.3D_k(U^\pi))\big]+G_k,\label{eq:loss}\\
 G_k={}&\tfrac{\kappa}{2}\left[\E\|\nabla_{U^E}D_k\|_2^p
                           +\E\|\nabla_{U^\pi}D_k\|_2^p\right],
 \quad \kappa=2,\ p=6.
 \label{eq:gp}
\end{align}
The penalty differentiates raw logits through the scan encoder with respect
to normalized windows, without expert--policy interpolation. One Adam step
minimizes $\sum_{k\in\mathcal A}\mathcal L_k$ over active classes, counting
shared parameters once. The current terrain head supplies the motion reward:
\begin{equation}
 \phi_k(U_t)=\exp\!\left(\tanh(0.3D_k(U_t))\right)-e^{-1}.
 \label{eq:style}
\end{equation}
\subsection{Support Candidate}
\label{sec:support_candidate}
We construct hand--knee support candidates for post-acquisition adaptation:
precontact guidance attracts approaching limbs, while actual-contact objectives
assess established support without prescribing a contact sequence.
Figure~\ref{fig:terrain} illustrates support geometry.

\input{figures/terrain_support}

\textbf{Surface candidates.}
A training-only map samples surfaces at $0.025$ m spacing over heights
$0.30$--$1.10$ m. Hand/knee footprints are $0.12\!\times\!0.08$/$0.10\!\times\!0.07$ m,
each tested at three yaw offsets. Supported-area thresholds are $0.85/0.92$,
normal-angle limits $0.35/0.25$ rad. The height standard deviation is limited to $0.015$ m. Score $s_G$ multiplies area, edge, and normal scores; edge score
saturates at $0.11$ m clearance. Nonmaximum suppression at $0.10$ m retains
at most 32 candidates per limb type and cell.

\textbf{Reachability and ranking.}
A body-aligned ellipsoid filters candidates relative to each limb's configured
anchor. For candidate $c$ and limb $\ell$,
\begin{equation}
 \begin{aligned}
 d_\ell(c)&=\sum_{j=1}^{3}\left(\frac{\bar c_j-\mu_{\ell,j}}{a_{\ell,j}}\right)^2,\\
 s_\ell(c)&=s_G(c)e^{-d_\ell(c)}\mathbf1[d_\ell(c)\leq1],
 \end{aligned}
 \label{eq:reach}
\end{equation}
where $\bar c$ is expressed in the aligned anchor frame, and $\mu_\ell$
and $a_\ell$ are the ellipsoid center and semiaxes. The hand center/axes
are $(.15,0,-.15)$ and $(.48,.42,.50)$ m; knee values are
$(.12,0,-.18)$ and $(.38,.285,.43)$ m. Up to eight candidates per limb remain.
The cache refreshes every five control steps and on reset or terrain-cell
change. The ellipsoid provides an approximate local reachability filter.

\textbf{Precontact guidance and actual support.}
For endpoint position $p_\ell$ and velocity $\dot p_\ell$, predict
$\hat p_\ell=p_\ell+\delta\dot p_\ell$ with $\delta=0.12$ s. The precontact
support score over the retained candidate set $\mathcal C_\ell$ is
\begin{equation}
 I_\ell=\sum_{c\in\mathcal C_\ell}
 \frac{e^{s_\ell(c)/T}}{\sum_{c'\in\mathcal C_\ell}e^{s_\ell(c')/T}}
 \exp\!\left[-\frac{\|\hat p_\ell-c\|^2}{\sigma_\ell^2}\right],
 \quad T=.15,
 \label{eq:imagined}
\end{equation}
with $\sigma_{\rm hand}=.12$ m and $\sigma_{\rm knee}=.10$ m.
For an empty candidate set, $I_\ell=0$.
Actual support $A_\ell$ uses the measured endpoint, a local surface mask,
$0.08$ m height tolerance, and 5/3 N contact hysteresis. The mask selects the
best sampled yaw, not the endpoint orientation. Its $0.5$ edge threshold
corresponds to approximately $0.055$ m clearance. Knee attraction and holding
require valid hand contact.

Before phase modifiers, the pair-averaged positive score is
\begin{equation}
 S_{\rm place}=.15\bar I_H+.35\bar A_H+.10\bar I_K+.25\bar A_K.
 \label{eq:place}
\end{equation}
Bars average left/right limbs ($H$: hands; $K$: knees). All four positive
terms require at least one cached reachable candidate across limbs. Actual
support is independent of the selected candidate position, not this availability
gate. Lateral safety logic uses actual support masks even with an empty
candidate set. Training gates do not define evaluation success.

\textbf{Support retention.}
For surfaces at least $0.90$ m high, define
$\rho=\max(|x_b-x_c|,|y_b-y_c|)$ from world-frame base and frame-center
coordinates. A lateral-support state latches when
$2.9\leq\rho<3.45$ m and the magnitude of the body-frame projected-gravity
$x$ component is at least $0.85$. Descent latches at $\rho\geq3.45$ m and
suppresses positive attraction and holding until reset. Low and Mid frames
remain in the mount state; these thresholds are not scaled to all heights.

Lateral holding requires at least two valid contacts including a hand;
two hands suffice. Holding/extra-contact weights are $0.15/0.025$. Penalties
for support deficit, no hand, contact loss, excessive swing, unprotected
release, simultaneous hand loss, and fast touchdown have respective weights
$0.35,0.35,0.10/0.05$ (hand/knee loss), $0.10,0.20,0.35,0.15$.
Hand and knee swing limits are 10 and 25 control steps. The full term is
$S=S_{\rm place}+S_{\rm hold}-P_{\rm support}$ with phase gates applied.
Other task rewards and geometric penalties remain outside $S$.

\textbf{Support-guided prior relaxation.}
During adaptation, geometric support gains emphasis while AMP guidance
is gradually reduced rather than removed. For control-step count $n$,
\begin{equation}
 \alpha(n)=0.1+0.9\clip\left(\frac{n}{96000},0,1\right).
 \label{eq:ramp}
\end{equation}
It reaches one after 4,000 updates of 24 control steps, scaling positive and
negative terms in $S$. AMP weight decreases via $\beta:1\rightarrow0.5$.

\section{Experiments}
\label{sec:evaluation}
We assess framework-level traversal (Sec.~\ref{sec:c1}), temporal
terrain--skill correspondence (Sec.~\ref{sec:c2}), support candidates and
prior relaxation (Sec.~\ref{sec:c3}), and zero-shot hardware traversal
(Sec.~\ref{sec:c4}). Unless explicitly
varied, comparisons share task rewards, reference data, and policy architecture;
stage-removal baselines do not receive the omitted stage's training budget.

All policies are trained in \emph{Isaac Lab}~\cite{nvidia_isaaclab_2025}
using an on-policy PPO implementation based on the RSL-RL training
stack~\cite{schulman_proximalpolicy_2017,schwarke_rslrllearning_2025}.
The deployment protocol additionally includes controller validation in
\emph{MuJoCo} and \emph{Gazebo} before hardware testing. Real-world evaluation
uses the \emph{Noetix E1}, a 23-DOF humanoid equipped with an onboard IMU,
joint encoders, and a base-mounted depth camera whose optical axis is pitched
by $60^{\circ}$ relative to the horizontal plane. The deployed policy runs
onboard an NVIDIA Jetson Orin NX and outputs joint position commands,
consistent with the actor's sensing and actuation interfaces.

\textbf{Training Details.}
The reference clips contain 335 standing, 785 turning, 735 lateral,
770 backward, 718 forward, 174 low-frame, 136 mid-frame, and 180 high-frame
frames, with $\Delta t=0.02$ s and common joint/body ordering.

Stage~1 reference preparation uses 20,000 training iterations;
Stage~2 skill acquisition uses 20,000 iterations, followed by 10,000
iterations of Stage~3 support adaptation. The three-stage schedule totals
50,000 iterations, of which 30,000 train the traversal policy in Stages~2
and~3. Table~\ref{tab:parameters} summarizes training settings.

\begin{table}[t]
\caption{Training configuration shared by the formal comparisons.}
\label{tab:parameters}\centering\small
\begin{tabularx}{\columnwidth}{@{}lY@{}}
\toprule
Parameter & Value\\
\midrule
Parallel environments / rollout & 4,096 / 24 control steps\\
PPO epochs / minibatches & 5 / 4\\
PPO learning rate / target KL & $10^{-3}$ adaptive / $0.01$\\
Clip / discount / GAE & $0.2$ / $0.99$ / $0.95$\\
Entropy / value loss / max. gradient & $0.01$ / $1$ / $1$\\
Discriminator learning rate & $5\times10^{-6}$\\
Adam decay: encoder/trunk; heads & $10^{-3}$; $0.1$\\
Replay capacity / history & 50,000 total / 4 frames\\
Terrain proportions: flat; each frame & $0.1$; $0.3$\\
Iterations: Stages~1 / 2 / 3 & 20,000 / 20,000 / 10,000\\
\bottomrule
\end{tabularx}
\end{table}

\subsection{Framework Comparison}
\label{sec:c1}
\textbf{Experimental Setup.}
We test whether the complete training pipeline supports individual and
continuous traversal and maintains performance under the tested
terrain-height shifts. Individual Low, Mid, and High frame
obstacles test skill execution, while a hybrid course containing all three
obstacles in sequence tests the combined ability to execute and connect
skills. An additional out-of-distribution (OOD) condition tests terrain
heights 1--3 cm outside the corresponding training ranges in
Table~\ref{tab:terrains}. Hybrid order, obstacle spacing, approach poses,
and travel direction are fixed in a scenario bank shared across methods.
Each hybrid rollout uses one policy without intermediate resets,
reference-state initialization, or external skill commands.

For each method, condition, and training seed, we evaluate one scored
episode in each of 2,048 parallel environments. Validation and test scenarios
are separate, and checkpoints are selected using validation performance
only. Single-obstacle and hybrid episodes have time limits of 20 and
60 s, respectively.

We measure complete-traversal success rather than conformity to a prescribed
contact pattern. Success requires traversing the designated obstacle region,
clearing its exit with the robot's collision envelope, and maintaining
stable support for one second after the final obstacle. Hybrid success
requires all three crossings in order, without mandatory stops between
them. Bypassing an obstacle, falling, requiring assistance, or timing out
counts as failure; no specific hand or knee contact is required.

\textbf{Methods.}
We compare the full framework (\textbf{Ours}) with three alternative
learning approaches and two stage-removal variants. \textbf{Velocity Tracking} uses task rewards without
an adversarial motion prior. \textbf{Single AMP} uses one terrain-independent
prior over pooled expert motions~\cite{peng_amp}, with the same total
expert-window budget as Ours. \textbf{Multi-Skill Distillation} is a local
baseline based on Perceptive Humanoid Parkour
(PHP)~\cite{wu_perceptivehumanoid_2026}: single-skill mimic policies first
learn individual reference motions, and their behaviors are then distilled
into one policy. Comparisons concern this local implementation.

The ablation variants examine the role of the progressive training
stages within the implemented schedule. \textbf{w/o Stage~2} omits
terrain-indexed skill acquisition, while \textbf{w/o Stage~3} evaluates
the policy before support adaptation. Neither variant reallocates the
omitted training updates. 

\textbf{Results and Analysis.}
As shown in Table~\ref{tab:main}, Ours achieves 100\%, 98.44\%, and 98.14\%
success on Low, Mid, and High, respectively, together with 96.34\% Hybrid
and 93.46\% OOD success. Velocity Tracking and Single AMP fail on High and
Hybrid. Multi-Skill Distillation matches Ours on Low and slightly outperforms
it on Mid, but falls behind on High, Hybrid, and OOD. These comparisons
highlight the full framework's advantage in demanding and continuous
traversal, beyond individual-skill performance, and under the tested height shifts.

Removing Stage~2 reduces success across all obstacle classes and yields
no successful hybrid traversals. Without Stage~3, High and Hybrid success
are 79.69\% and 83.59\%. These comparisons favor the complete training
schedule. Because omitted stages also remove training updates, the results
do not isolate each stage's objective from its additional training budget.
Contact-safety conclusions require measurements beyond traversal success.

\begin{table}[t]
\caption{Complete-traversal success (\%): 2,048 episodes per condition. Low/Mid/High follow Table~\ref{tab:terrains}. Hybrid requires
all three obstacles without resets; OOD heights lie 1--3 cm outside the
corresponding training ranges.}
\label{tab:main}\centering\small
\setlength{\tabcolsep}{5pt}
\begin{tabular}{@{}lccccc@{}}
\toprule
Method & Low & Mid & High & Hybrid & OOD\\
\midrule
Ours & \textbf{100} & 98.44 & \textbf{98.14} & \textbf{96.34} & \textbf{93.46} \\
w/o Stage~2 & 41.36 & 21.14 & 13.87 & 0 & 0 \\
w/o Stage~3 & \textbf{100} & 85.55 & 79.69 & 83.59 & 85.11 \\
Velocity Tracking & 96.09 & 0 & 0 & 0 & 0 \\
Single AMP & \textbf{100} & 65.82 & 0 & 0 & 44.29 \\
Multi-Skill Distillation & \textbf{100} & \textbf{99.28} & 95.10 & 79.65 & 65.33 \\
\bottomrule
\end{tabular}
\end{table}

\subsection{Temporal Terrain--Skill Correspondence}
\label{sec:c2}
\textbf{Experimental Setup.}
To test whether reference similarity changes during a successful traversal,
we construct a row
of three elevated platforms in Isaac Sim and load the target policy.
Each platform is $0.5$ m thick along the forward ($X$) direction, with
$4$ m clear gaps between adjacent platforms. The robot starts $1.5$ m
before the leading edge of the first platform. Reference-state
initialization (RSI) and the curriculum are disabled during collection,
and the commanded forward linear velocity is $1.5$ m/s.

A valid crossing must approach the target platform from the front and
pass both its leading and trailing edges. While the root lies within
the platform's $X$ extent, its $Y$ coordinate must remain within
$[4.7,7.3]$ m, leaving a $0.2$ m margin on each side to exclude bypassing.
A crossing is successful once the root moves at least $0.15$ m beyond
the trailing edge and remains alive behind the platform for $0.5$ s
continuously. Falling, timing out, a manual reset, or exiting terminates
the current attempt. An attempt that has not completed the traversal
within $20$ s of motion onset is counted as a failure.

\begin{figure}[t]
\centering
\includegraphics[width=\columnwidth]{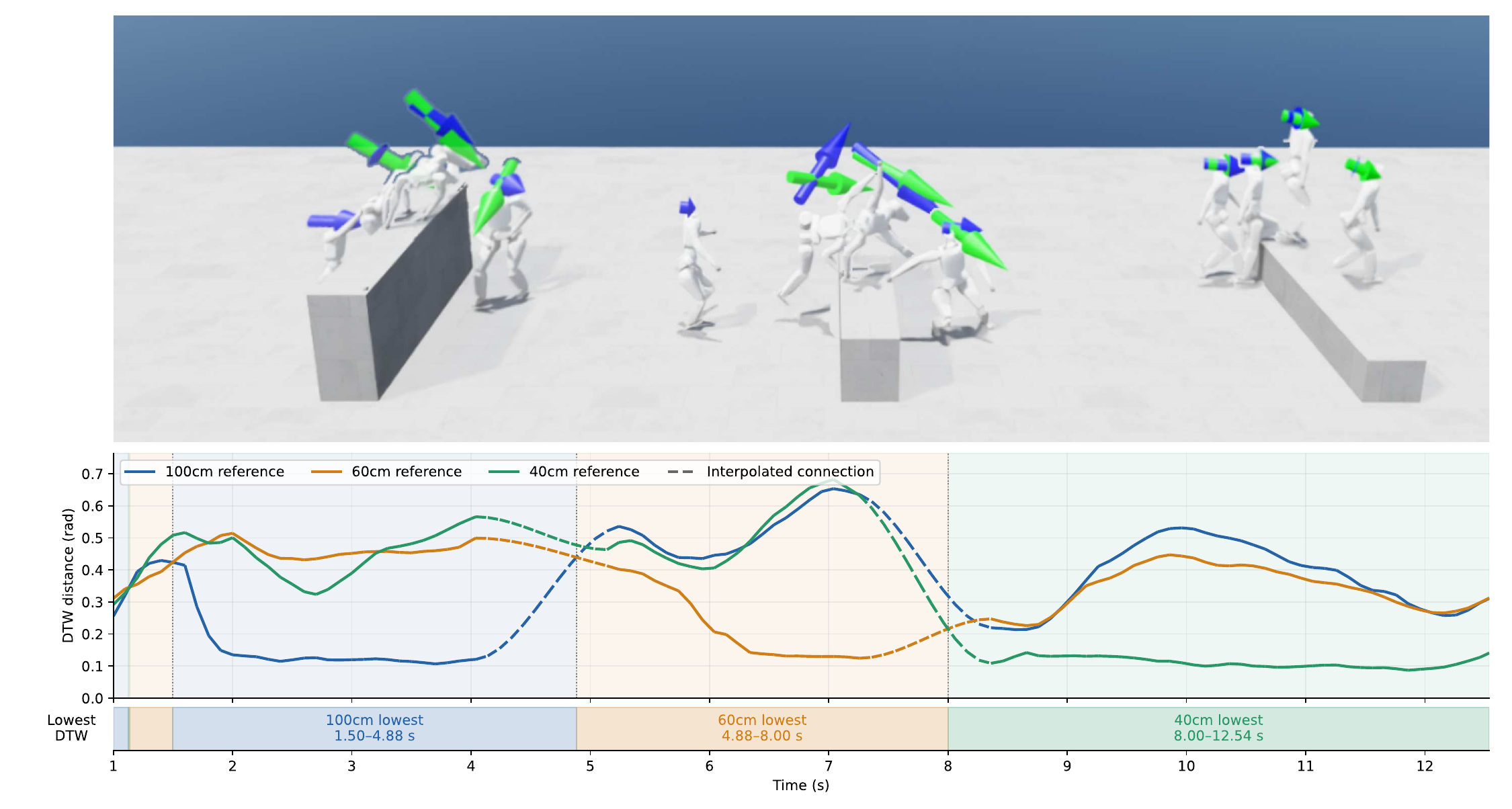}
\caption{Supplementary evidence for terrain--skill correspondence.
Top: a simulation montage of sequential obstacle traversal. Middle: joint-angle
DTW distances (rad) to the High, Mid, and Low references. Bottom: lowest-distance
intervals, with the dominant sequence High$\rightarrow$Mid$\rightarrow$Low.
The interval labels identify reference matches rather than independently
measured terrain boundaries.}
\label{fig:temporal_protocol}
\end{figure}

We retain a successful continuous traversal segment from a single attempt
and record all 23 measured joint angles at every sampled time step.
Segments from different attempts are never concatenated.
Figure~\ref{fig:temporal_protocol} compares this trajectory with High,
Mid, and Low motion references as a qualitative analysis of motion changes,
not an estimate of traversal success rate.

We use dynamic time warping (DTW), motivated by
HIL~\cite{wang_hil_2025}, with the following distance definition.
Inputs are the robot's executed 23-dimensional absolute joint angles and
the standard E1 references for 100, 60, and 40 cm obstacles (High, Mid,
and Low). Joints are aligned by name, expressed in radians, and weighted
equally, without amplitude normalization. Position, velocity, scans, and
terrain labels are excluded. For sequences of lengths $N$ and $M$, the
framewise cost is
\begin{equation}
 c(i,j)=\frac{1}{23}\sum_{k=1}^{23}
 (q_{i,k}-q^{\rm ref}_{j,k})^2.
 \label{eq:dtw_cost}
\end{equation}
The symmetric2 recurrence is
\begin{equation}
 D(i,j)=\min\!\begin{cases}
 D(i-1,j-1)+2c(i,j),\\
 D(i-1,j)+c(i,j),\\
 D(i,j-1)+c(i,j),
 \end{cases}
 \label{eq:dtw_recurrence}
\end{equation}
initialized with $D(0,0)=2c(0,0)$ and using valid predecessors at
sequence boundaries. The final distance is
\begin{equation}
 d=\sqrt{\frac{D(N-1,M-1)}{N+M}}.
 \label{eq:dtw_distance}
\end{equation}
Normalization uses the sum of sequence lengths, not the number of path
nodes. The distance is in radians; smaller values indicate more similar
joint motion.

\textbf{Results and Analysis.}
Figure~\ref{fig:temporal_protocol} shows the closest reference progressing
from High ($1.50$--$4.88$ s) to Mid ($4.88$--$8.00$ s) and Low
($8.00$--$12.54$ s).
Together with the traversal montage, this provides qualitative evidence
that a single perception-conditioned policy expresses different
terrain-associated skill patterns without external skill-selection commands.
The selected successful trajectory does not estimate switching accuracy
or distinguish learned transitions from motion similarity along this route.

\subsection{Candidate-Guided Support Adaptation}
\label{sec:c3}
\textbf{Experimental Setup.}
We test whether the complete support-adaptation stage improves traversal
on the Low, Mid, and High frames from Sec.~\ref{sec:c1}.
To further assess geometric support guidance under a more challenging
support layout, we additionally test a 1-meter-high platform with a
$1\,\mathrm{m}\times1\,\mathrm{m}$ top surface.
This terrain requires the robot to maintain support on the planar top
while moving laterally across it, increasing the risk of falling from
an edge when support is lost or misplaced.
It therefore provides a more demanding test of support maintenance and
fall avoidance, assessed through complete-traversal success.
Each condition follows the 2,048-environment and
complete-traversal criterion in Sec.~\ref{sec:c1}.

\textbf{Methods.}
Table~\ref{tab:candidate} compares four variants starting from the same
Stage~2 checkpoint. \textbf{Ours} uses both support-candidate guidance
and prior relaxation. \textbf{The w/o Candidates} variant removes the reward terms associated with support candidates
while retaining prior relaxation. \textbf{The w/o Relaxation} variant retains support candidates but fixes
AMP weight at its Stage~2 value. \textbf{The w/o Stage~3} variant evaluates the
Stage~2 checkpoint without support adaptation.

\begin{figure}[t]
\centering
\includegraphics[width=\columnwidth,trim=125bp 145bp 65bp 0bp,clip]{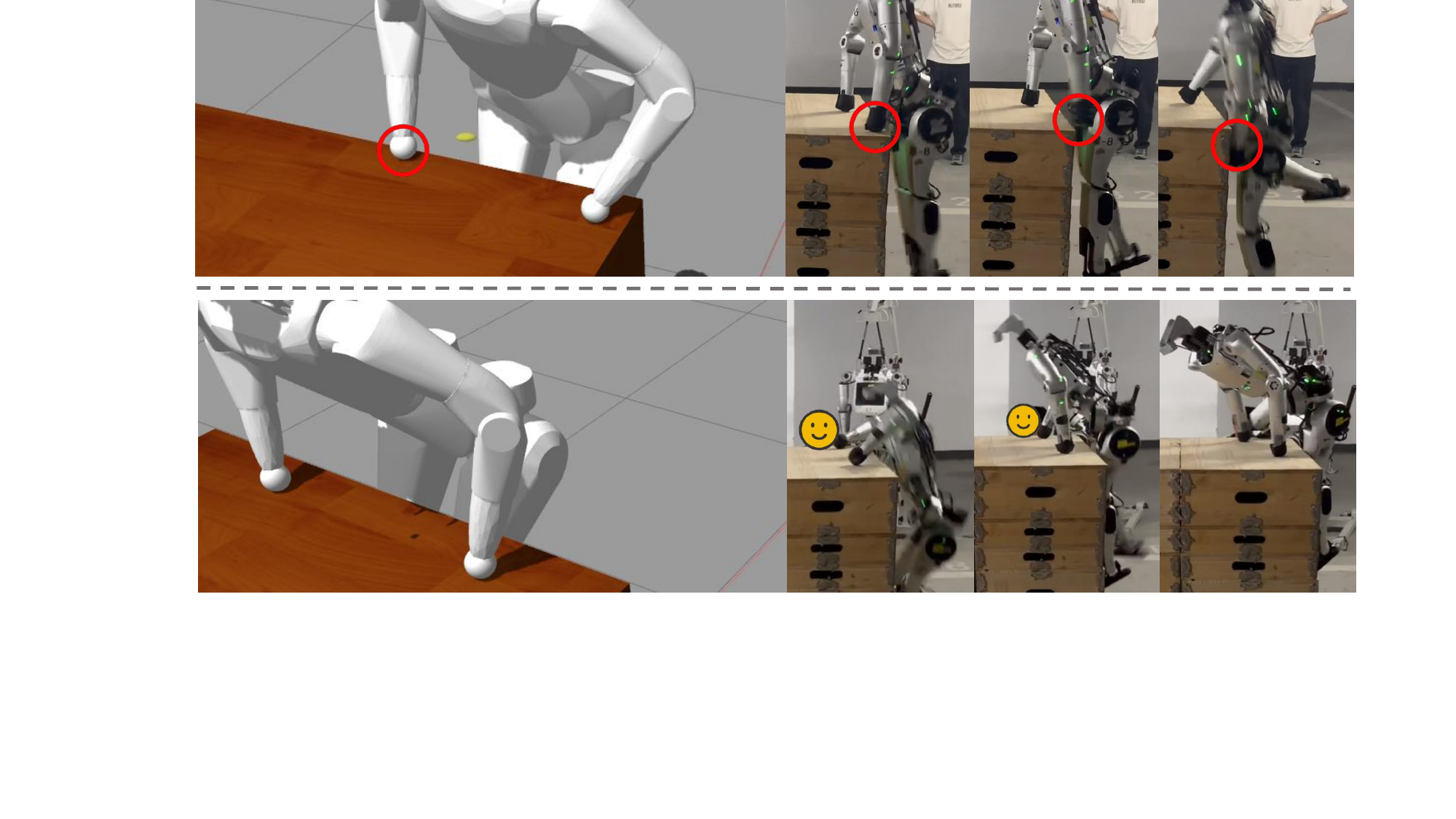}
\caption{Qualitative support comparison. Top: w/o Stage~3;
bottom: Ours. Each row combines a simulation contact close-up
(left) and hardware traversal snapshots (right). Red circles in the upper
row highlight hand-contact regions near the platform edge. The comparison
illustrates support adaptation rather than a measured contact-force or
stability margin.}
\label{fig:stage3_support}
\end{figure}

\begin{table}[htb]
\caption{Success rates for support-adaptation ablations (\%).}
\label{tab:candidate}\centering\small
\begin{tabular}{lcccc}
\toprule
Method & Low & Mid & High & $1\,\mathrm{m}\!\times\!1\,\mathrm{m}$ Top\\
\midrule
Ours & \textbf{100} & \textbf{98.44} & \textbf{98.14} & \textbf{93.46} \\
w/o Candidates & 99.17 & 83.15 & 71.53 & 46.24 \\
w/o Relaxation & 99.56 & 85.40 & 83.20 & 58.89 \\
w/o Stage~3 & \textbf{100} & 85.55 & 79.69 & 50.63 \\
\bottomrule
\end{tabular}
\end{table}

\textbf{Results and Analysis.}
Table~\ref{tab:candidate} shows little separation on Low, but a clearer
advantage for the full method on Mid, High, and especially the platform
requiring sustained lateral support.
On this platform, Ours achieves 93.46\% success, compared with 46.24\%
without support-candidate rewards and 58.89\% without prior relaxation, supporting the
combined design for traversal on a limited support surface.
The full method also outperforms the pre-adaptation checkpoint, consistent
with the benefit of the complete support-adaptation stage.
The reward ablation evaluates the support-candidate reward terms together.
The additional Stage~3 training limits attribution of the comparison with
the pre-adaptation checkpoint to adaptation objectives alone.

Figure~\ref{fig:stage3_support} complements the success rates with qualitative
contact observations: the pre-adaptation examples show edge-adjacent hand
contacts, whereas the full method exhibits a different support configuration.
Together, the results support improved traversal performance and changes
in support placement under the complete adaptation stage, particularly on
the additional platform, rather than directly establishing lower contact
forces, larger stability margins, or a general contact-safety guarantee.

\subsection{Real-World Single and Hybrid Traversal}
\label{sec:c4}
\textbf{Experimental Setup.}
We test whether the simulation-trained policy transfers to onboard hardware
control without real-world fine-tuning. We evaluate individual Low, Mid,
and High obstacles in ten trials each,
conduct five continuous hybrid trials without intermediate resets,
and perform 15 additional OOD trials. The OOD condition follows
Sec.~\ref{sec:c1}, with obstacle heights 1--3 cm outside the corresponding
training ranges in Table~\ref{tab:terrains}.
Success requires complete traversal and a stable final landing
without assistance. The 30 single-obstacle, five hybrid, and 15 OOD trials
are counted separately; intermediate crossings within a hybrid course
are not treated as separate trials.

\textbf{Methods.}
The same simulation-trained policy is deployed zero-shot on the Noetix E1
across all three settings, without real-world policy fine-tuning. This tests
transfer under combined actuator, contact, and perception differences,
without separately measuring these sources of mismatch. It runs onboard
the NVIDIA Jetson Orin NX using onboard perception and proprioception. Both the individual obstacle
behaviors and transitions between terrain types are executed by one policy.

\begin{figure}[t]
\centering
\includegraphics[width=\columnwidth,trim=195bp 215bp 410bp 181bp,clip]{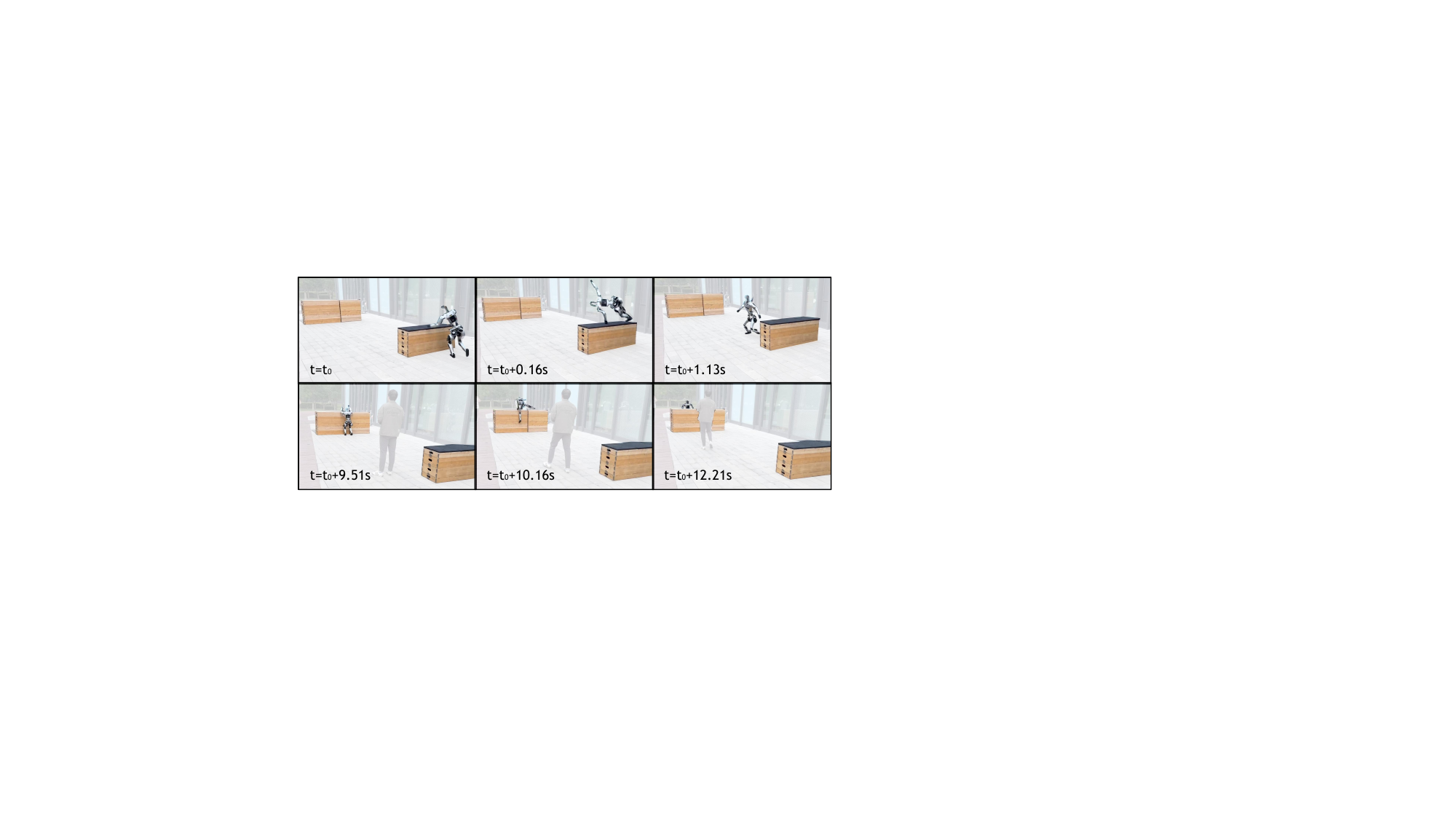}
\caption{Supplementary real-world hybrid traversal sequence. Snapshots are
ordered from left to right, top to bottom, with timestamps relative to $t_0$.
The two rows show different obstacle-crossing maneuvers within the same trial,
illustrating terrain-dependent skill switching by a single perception-conditioned
policy without external skill-selection commands.}
\label{fig:hardware_hybrid}
\end{figure}

\begin{table}[htb]
\caption{Real-world traversal results (successes/attempts).}
\label{tab:hardware}\centering\small
\begin{tabular}{lccccc}
\toprule
Policy & Low & Mid & High & Hybrid & OOD\\
\midrule
Ours & 10/10 & 9/10 & 10/10 & 4/5 & 13/15\\
\bottomrule
\end{tabular}
\end{table}

\textbf{Results and Analysis.}
Table~\ref{tab:hardware} reports 10/10, 9/10, and 10/10 successes
on Low, Mid, and High, respectively, giving 29/30 successful
single-obstacle trials. The same zero-shot policy completes 4/5
hybrid courses and succeeds in 13/15 OOD trials, supporting
continuous traversal and generalization to the tested terrain heights
outside the training ranges.

The Mid failure involves loss of arm support. In the unsuccessful
hybrid trial, an abrupt heading change precedes a cartwheel-style
maneuver associated with Mid obstacles on a Low platform and traversal
failure. These observations identify support maintenance and
terrain-appropriate skill transitions as remaining challenges.

Figure~\ref{fig:hardware_hybrid} provides complementary qualitative
evidence of skill switching: the same policy adopts different
traversal behaviors across successive obstacles using onboard
perception and proprioception, without intermediate resets or
external skill-selection commands.

\section{Conclusion}
We presented a progressive imitation framework that uses terrain-indexed
motion priors to acquire humanoid traversal skills, then relaxes these
priors and introduces geometric hand--knee support guidance.
Support-adaptation ablations show higher traversal success for the full
adaptation stage on the tested terrains, while qualitative observations
show changes in contact placement.
The resulting single policy achieves 96.34\% hybrid success in simulation
and completes 29/30 individual obstacle trials and 4/5 hybrid courses on
hardware without real-world fine-tuning.
These results support first acquiring skills from demonstrations, then
relaxing their influence to accommodate robot-specific support needs.
Future work will broaden reference coverage and develop support-candidate
generation for irregular surfaces.

\balance

\bibliographystyle{IEEEtran}
\bibliography{references}
\end{document}

%% file: figures/framework.tex
\begin{figure*}[t]
\centering
\includegraphics[width=\textwidth]{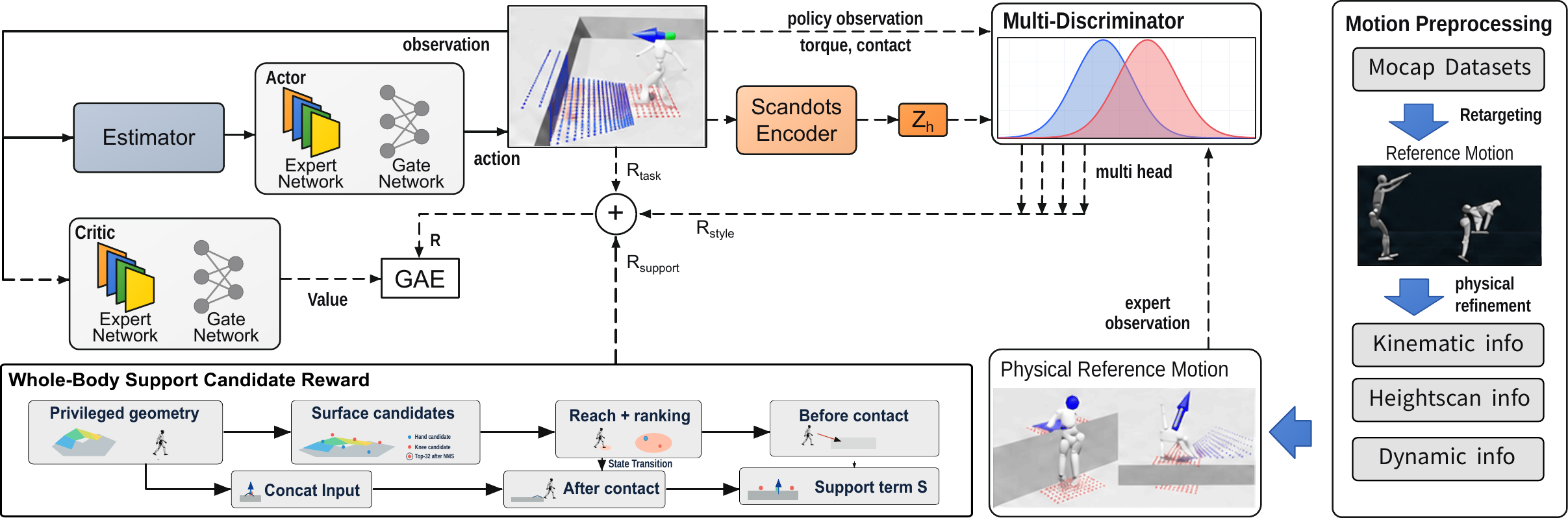}
\caption{Overview of the training framework. Motion preprocessing (right)
retargets human demonstrations and physically refines them into robot reference
rollouts with kinematic, height-scan, and dynamic information. Terrain-indexed
adversarial supervision compares reference and policy motion windows to provide
a routed style reward. Geometric hand--knee candidates (bottom) guide precontact
placement, while contact evaluation and state transitions determine support
and retention terms. Task, style, and support rewards are combined with the
weights in Eq.~\ref{eq:reward} for actor--critic training. The actor and critic
use gated expert networks, distinct from the discriminator heads. At deployment,
only the actor and its observation-processing modules are retained; privileged
geometry, reference motions, discriminators, and the critic remain in training.}
\label{fig:pipeline}
\end{figure*}

%% file: figures/terrain_support.tex
\begin{figure}[t]
\centering
\resizebox{\columnwidth}{!}{\input{figures/support_candidate_diagram}}
\caption{Support Candidate mechanism (schematic, not to scale).
(a) The frame's open center is excluded from support surfaces.
(b) Footprint compatibility and approximate 3-D ellipsoidal reachability filter
candidate sets; the dashed ellipse illustrates a projection. Filled circles
with rectangles denote valid candidate footprints, a cross denotes edge
rejection, and an open circle denotes an unreachable candidate.
(c) Predicted endpoints receive weighted attraction to valid candidates.
(d) Actual contact quality is queried at the measured endpoint, with contact,
candidate-availability and phase gates. The requested band width is $0.55$ m;
the native horizontal top is approximately $0.45$ m wide. }
\label{fig:terrain}
\end{figure}

%% file: figures/support_candidate_diagram.tex
\definecolor{scblue}{HTML}{0077BB}
\definecolor{scteal}{HTML}{009988}
\definecolor{scorange}{HTML}{CC7733}
\definecolor{scpurple}{HTML}{AA3377}
\begin{tikzpicture}[x=1cm,y=1cm,>=Latex,font=\fontsize{8}{9.2}\selectfont,
 point/.style={circle,draw=scteal,fill=scteal!20,inner sep=1.8pt},
 lab/.style={fill=white,inner sep=1pt,font=\fontsize{8}{9.2}\selectfont}]
\node[anchor=west,font=\fontsize{9}{10}\selectfont\bfseries] at (0,6.8) {(a) Frame top view};
\node[anchor=west,font=\fontsize{9}{10}\selectfont\bfseries] at (4.1,6.8) {(b) Candidate filtering};
\fill[scorange!15,even odd rule] (.2,4.0) rectangle (2.75,6.55)
 (.72,4.52) rectangle (2.23,6.03);
\draw[scorange,thick] (.2,4.0) rectangle (2.75,6.55);
\draw[scorange,thick] (.72,4.52) rectangle (2.23,6.03);
\node[align=center] at (1.475,5.275) {Open center\\(ground)};
\draw[->,thick,scblue] (1.55,4.9)--(3.35,4.9);
\node[anchor=north,align=center] at (1.55,3.86) {Real top band; hole excluded};
\draw[dashed,black!60] (2.12,4.6) rectangle (2.87,5.78);
\draw[dashed,black!45] (2.87,5.78)--(4.22,6.37);
\draw[dashed,black!45] (2.87,4.6)--(4.22,4.3);

\fill[black!4] (4.22,4.3) rectangle (5.0,6.37);
\fill[scorange!15] (5.0,4.3) rectangle (8.25,6.37);
\draw[scorange,thick] (5.0,4.3)--(5.0,6.37);
\draw[scblue,dashed,thick] (5.44,5.3) ellipse (1.55 and .82);
\fill[scblue] (4.43,5.3) circle (.045);
\node[lab,anchor=east] at (4.88,4.62) {Anchor};
\draw[->,scblue] (4.63,4.79)--(4.43,5.23);
\node[point] at (5.6,5.62) {};
\draw[scteal] (5.4,5.48) rectangle (5.8,5.76);
\node[point] at (6.26,5.3) {};
\draw[scteal] (6.06,5.16) rectangle (6.46,5.44);
\node[point] at (5.85,4.93) {};
\draw[scteal] (5.65,4.79) rectangle (6.05,5.07);
\draw[scorange,thick] (7.61,5.9) circle (.08);
\node[lab,align=center] at (7.55,4.68) {Outside\\reach};
\draw[->,scorange] (7.55,5.05)--(7.61,5.78);
\draw[scpurple] (4.75,5.85) rectangle (5.15,6.13);
\draw[scpurple,thick] (4.88,5.92)--(5.03,6.07) (4.88,6.07)--(5.03,5.92);
\node[lab,anchor=west] at (5.24,6.18) {Footprint crosses edge};
\node[align=center,anchor=north,text width=4.05cm] at (6.23,3.86)
 {Hand / knee surface footprints\\$+$ approximate reach filter};

\node[anchor=west,font=\fontsize{9}{10}\selectfont\bfseries] at (0,2.91) {(c) Before contact};
\node[anchor=west,font=\fontsize{9}{10}\selectfont\bfseries] at (4.1,2.91) {(d) Actual contact};
\fill[scorange!12] (2.05,1.05) rectangle (3.65,2.54);
\node[point] (c1) at (2.55,2.23) {};
\node[point] (c2) at (3.15,1.88) {};
\node[point] (c3) at (2.58,1.39) {};
\fill[scblue] (.98,1.86) circle (.055);
\node[anchor=east,align=center,text=scblue] at (.88,1.88) {$\hat p$};
\draw[->,scblue] (.29,1.14)--(.93,1.8);
\node[anchor=west] at (0,1.0) {$p+0.12\dot p$};
\draw[->,dashed,scblue] (1.04,1.9)--(c1);
\draw[->,dashed,scblue,thick] (1.04,1.86)--(c2);
\draw[->,dashed,scblue] (1.04,1.82)--(c3);
\node[anchor=north,align=center,text width=3.8cm] at (1.83,.72)
 {Score-weighted attraction\\over valid candidates};

\fill[scorange!15] (4.35,1.08) rectangle (8.15,1.46);
\draw[scorange,thick] (4.35,1.46)--(8.15,1.46);
\draw[black!65,line width=1.3pt] (4.73,2.49)--(5.45,2.09)--(6.14,1.6);
\draw[scpurple,fill=scpurple!18,thick] (5.88,1.48) rectangle (6.4,1.76);
\fill[scpurple] (6.14,1.62)--(6.22,1.54)--(6.14,1.46)--(6.06,1.54)--cycle;
\draw[->,scpurple,thick] (6.14,1.77)--(6.14,2.48);
\node[anchor=west,text=scpurple] at (6.3,2.27) {Contact force};
\node[anchor=north,align=center,text width=3.9cm] at (6.2,.72)
 {Actual endpoint quality\\Force / cache / phase gates};
\draw[black!20] (0,-.02)--(8.25,-.02);
\node[anchor=north,align=center,text width=8.2cm] at (4.12,-.16)
 {\textbf{High only:} retain support $\to$ hand transfer $\to$ release\\
 Training schematic; no measured trajectory.};
\end{tikzpicture}